\pdfoutput=1
\documentclass[10pt,twocolumn,letterpaper]{article}

\usepackage{cvpr}
\usepackage{times}
\usepackage{epsfig}
\usepackage{graphicx}
\usepackage{subcaption}
\usepackage{amsmath}
\usepackage{amssymb}

\usepackage[breaklinks=true,bookmarks=false]{hyperref}

\cvprfinalcopy 

\def\cvprPaperID{****} 

\begin{document}

\title{Semi-Supervised Noisy Student Pre-training on EfficientNet Architectures for Plant Pathology Classification}

\author{Sedrick Scott Keh\\
The Hong Kong University of Science and Technology\\
\\
{\tt\small sskeh@connect.ust.hk}}
\maketitle

\begin{abstract}
In recent years, deep learning has vastly improved the identification and diagnosis of various diseases in plants. In this report, we investigate the problem of pathology classification using images of a single leaf. We explore the use of standard benchmark models such as VGG16, ResNet101, and DenseNet 161 to achieve a 0.945 score on the task. Furthermore, we explore the use of the newer EfficientNet model, improving the accuracy to 0.962. Finally, we introduce the state-of-the-art idea of semi-supervised Noisy Student training to the EfficientNet, resulting in significant improvements in both accuracy and convergence rate. The final ensembled Noisy Student model performs very well on this particular classification task, achieving a test score of 0.982.
\end{abstract}

\section{Introduction}
Plant pathology classification is a very challenging and important task. In 2019, more than 20$\%$ of all global crop productions were lost due to pests and other pathogens. This results in massive economic losses and food deficits in many parts of the world. Therefore, it is very important to be able to develop systems that can detect such diseases. Such systems will have a significant impact on the agricultural sector, especially in areas where agriculture is a primary food source for millions of people.

\begin{figure}[h!]
  \centering
  \begin{subfigure}[b]{0.9\linewidth}
    \includegraphics[width=\linewidth]{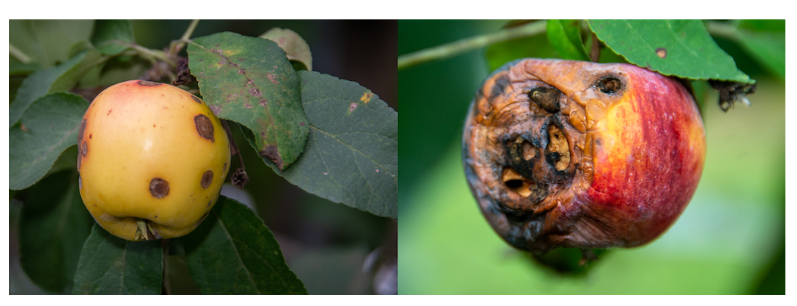}
  \end{subfigure}
  \caption{Examples of fungal diseases in apples \cite{dataset}}
  \label{fig:coffee}
\end{figure}

Currently, plant disease diagnosis is mostly done manually by human examination. However, this is inefficient, as it is tedious, time-consuming, and error-prone. Recently, deep learning and computer vision techniques have shown great promise in this task. These systems have the ability to quickly and reliably make predictions, and have greatly contributed to plant classification in the past few years. However, there are still many challenges. For instance, images of different leaves may look very similar with each other. Meanwhile, two images from the same leaf species may have very different images. Plant disease symptoms may also manifest in different ways due to age of infected tissues, genetic variations, and light conditions.

In this report, we examine the use of pretrained deep learning models in plant pathology classification. More specifically, we examine the state-of-the-art EfficientNet \cite{efficientnet} and a semi-supervised approach called \textbf{Noisy Student training} \cite{NoisyStudent}. We demonstrate that the Noisy Student EfficientNet performs very well on our dataset, and it has significant improvements over previous attempts at the task.
\section{Related Work}
\subsection{Traditional Methods}
Prior to the introduction of deep learning methods, plant pathology detection was mostly done through traditional image processing and classification techniques \cite{Survey}. Until now, some of these traditional methods remain widely used due to their simplicity.
\subsubsection{Image Processing Methods}
An image of a leaf can reveal many characteristics of a particular plant, such as color, style, and shape. Therefore, many early attempts at this task were focused on extracting features from these leaf characteristics. Using image processing techniques such as edge detectors to analyze image gradients, one can easily identify features such as discolorations, spots, or holes on the leaf surfaces \cite{Radha}.

While this idea makes intuitive sense, it may not work that well in all situations. This is because certain diseases may not be so easily visible from the image due to variations in lighting, obstruction, and position. Conversely, the discoloration seen in an image may not necessarily be from diseases as well. It is possible that these changes in color are just due to sunlight patterns or other external factors.

\subsubsection{Clustering}
There have also been various attempts to use clustering algorithms to group similar leaves together. The biggest advantage of using a clustering algorithm is that it is sensitive to patterns in local areas \cite{Clustering}. For example, some diseases may be focused more on damaging the edges of a leaf, while others may cause discolorations in the center of the leaf. The k-means clustering algorithm will be able to capture these relationships since it will group together leaves which have symptoms in the same general area.  \cite{clustering2}

However, one disadvantage of clustering is that once again, it may fail to pick up on certain patterns or be sensitive false patterns that are not actually there. Because the distance metric is usually based on pixel vales or other pixel features, it is prone to errors from variations in position, occlusion, lighting, and many others.

\subsubsection{SVM and Other Classification Models}
To address the issues of the previous methodologies, many recent approaches have explored the use of machine learning architectures \cite{Khirade}. For instance, a report by Singh \cite{Singh} explored the use of genetic algorithms in disease image classification. In their report, they first performed image processing to find certain features of the leaves, followed by a genetic optimization algorithm to reach a classification accuracy of $95.71\%$. Another report by Rumpf \cite{rumpf} uses SVMs and achieves a similar accuracy of $96\%$. These, accuracies, however, are only binary predictions, and fail to identify specific diseases or whether the plant has more than one disease. When predicting multiple diseases, the SVM accuracy drops significantly to around $86\%$.

\subsection{Deep Learning Based Methods}
More recently, deep neural networks and CNN-based models have greatly aided in plant disease classification \cite{Toda}, reaching accuracies as high as $97.4\%$ on some datasets. Furthermore, these also offer the advantage that there is no need to for manual feature extraction and engineering. Rather, these deep learning models can take in the pure images as training data and make predictions from them. This makes the process more scalable.
\subsubsection{Transfer Learning on Plant Classification} There have been a few studies on using transfer learning for plant disease classification. However, many of these studies are too highly specific and may not generalize well. For instance, one study \cite{rice} focused specifically on just rice crops, and achieved a score of around $91.37\%$ with a pre-trianed AlexNet model. 

Additionally, a report by Aravind $\&$ Raja \cite{good} explores the use of transfer learning models such as VGG16, VGG19, GoogLeNet, ResNet101 and DenseNet201, achieving an accuracy of $97.3\%$ on four different species of plants.  However, the state-of-the-art has become much better since then, with new models such as EfficientNet coming out. This means that the accuracy on the task can likely be improved. Furthermore, the dataset used was also quite limited. We aim to improve on this accuracy using more novel architectures and other model training strategies. 

\section{Data}
\subsection{Dataset}
The dataset used in this report is the same dataset from the Plant Pathology FGVC7 workshop of CVPR 2020 \cite{dataset}. The data was compiled by the Cornell Initiative for Digital Agriculture (CIDA), and it contains image data of 1820 apple leaves, together with corresponding labels of their diseases. Each leaf has 4 columns of binary values (“healthy”, “multiple$\_$diseases”, “rust”, “scab”). There are also 1820 testing images with no labels.

Four sample images are shown below
\begin{figure}[h!]
  \centering
  \begin{subfigure}[b]{0.4\linewidth}
    \includegraphics[width=\linewidth]{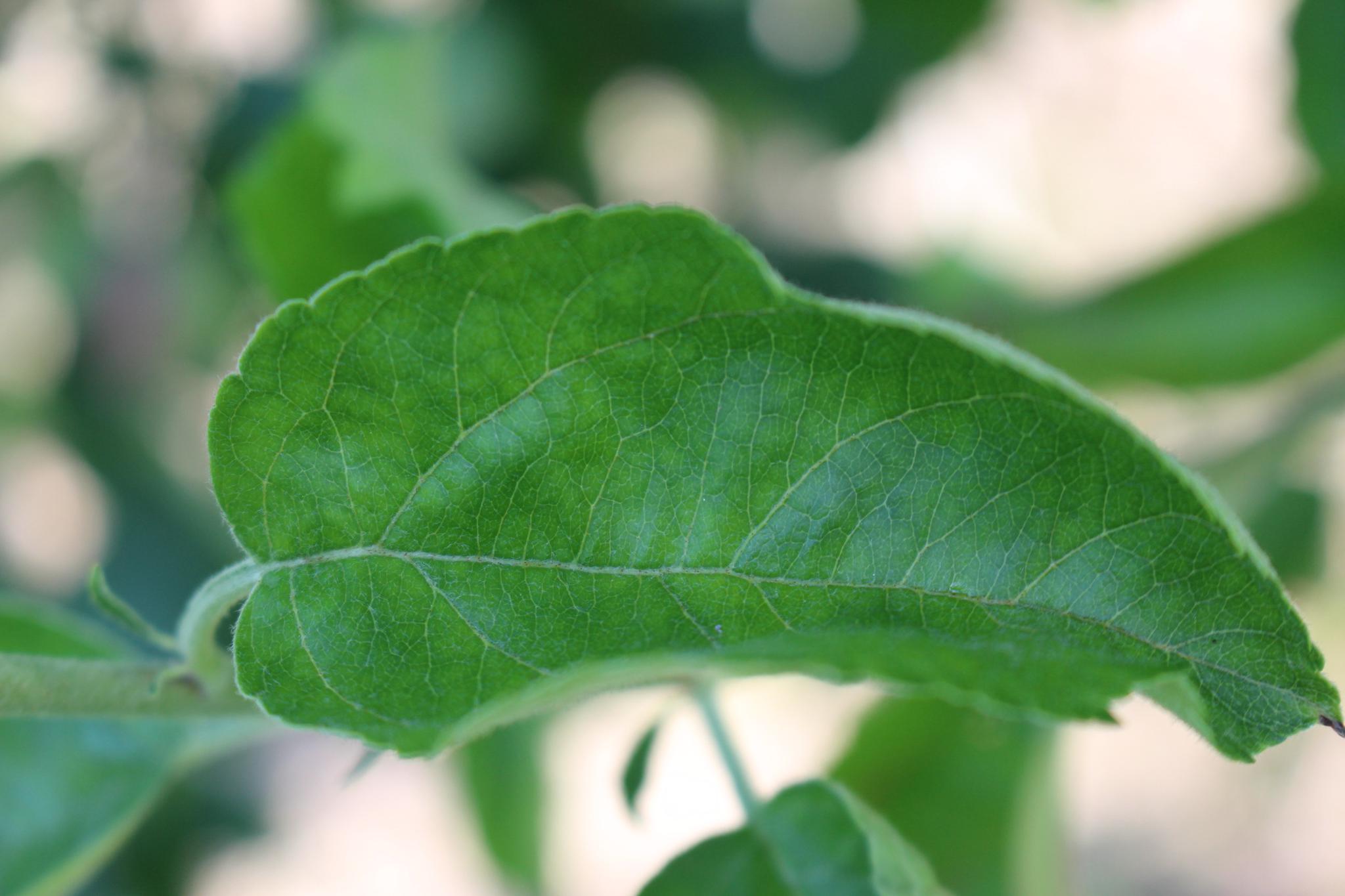}
    \caption{Healthy Leaf}
  \end{subfigure}
  \begin{subfigure}[b]{0.4\linewidth}
    \includegraphics[width=\linewidth]{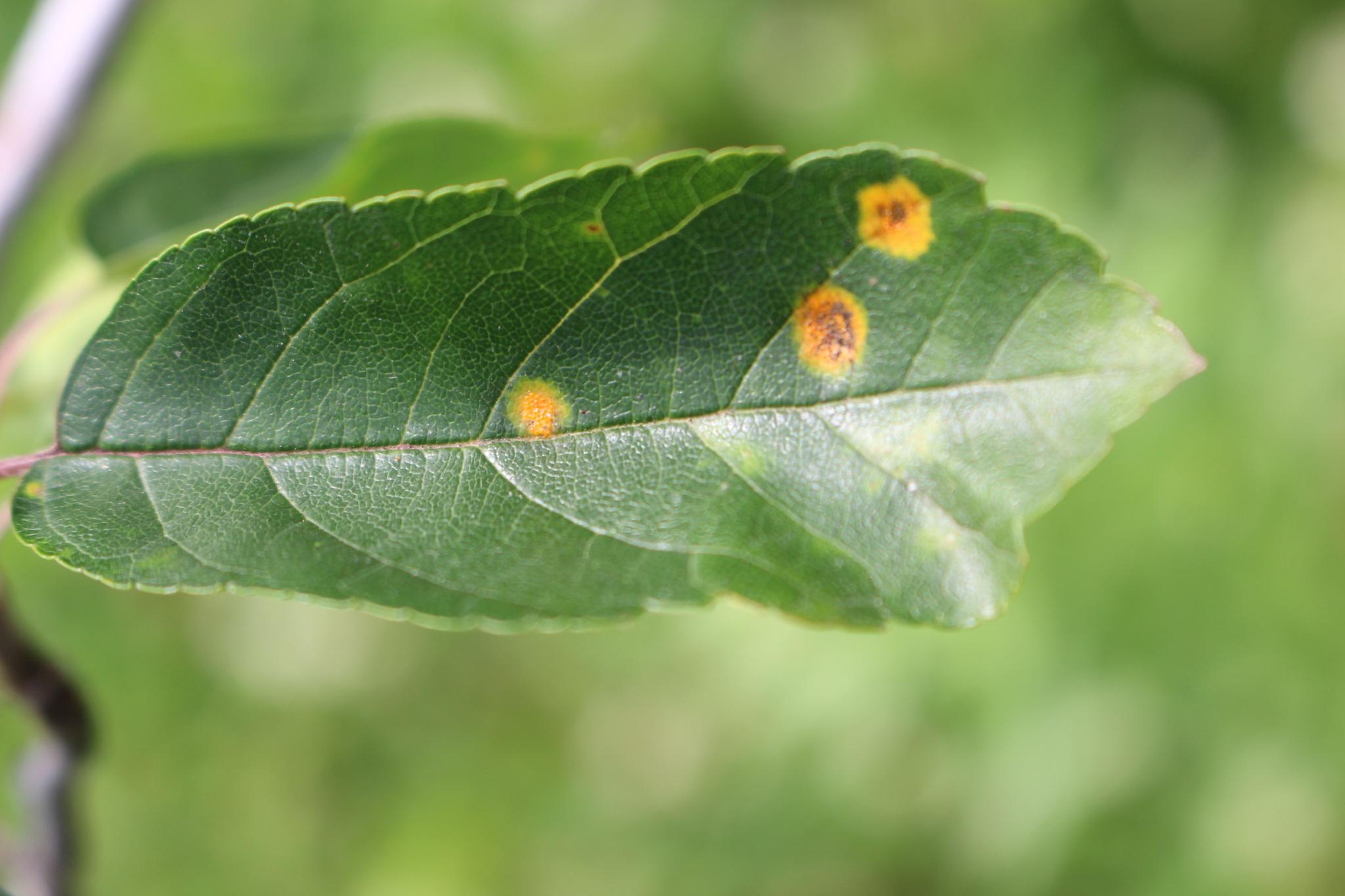}
    \caption{"Rust" Disease}
  \end{subfigure}
  \begin{subfigure}[b]{0.4\linewidth}
    \includegraphics[width=\linewidth]{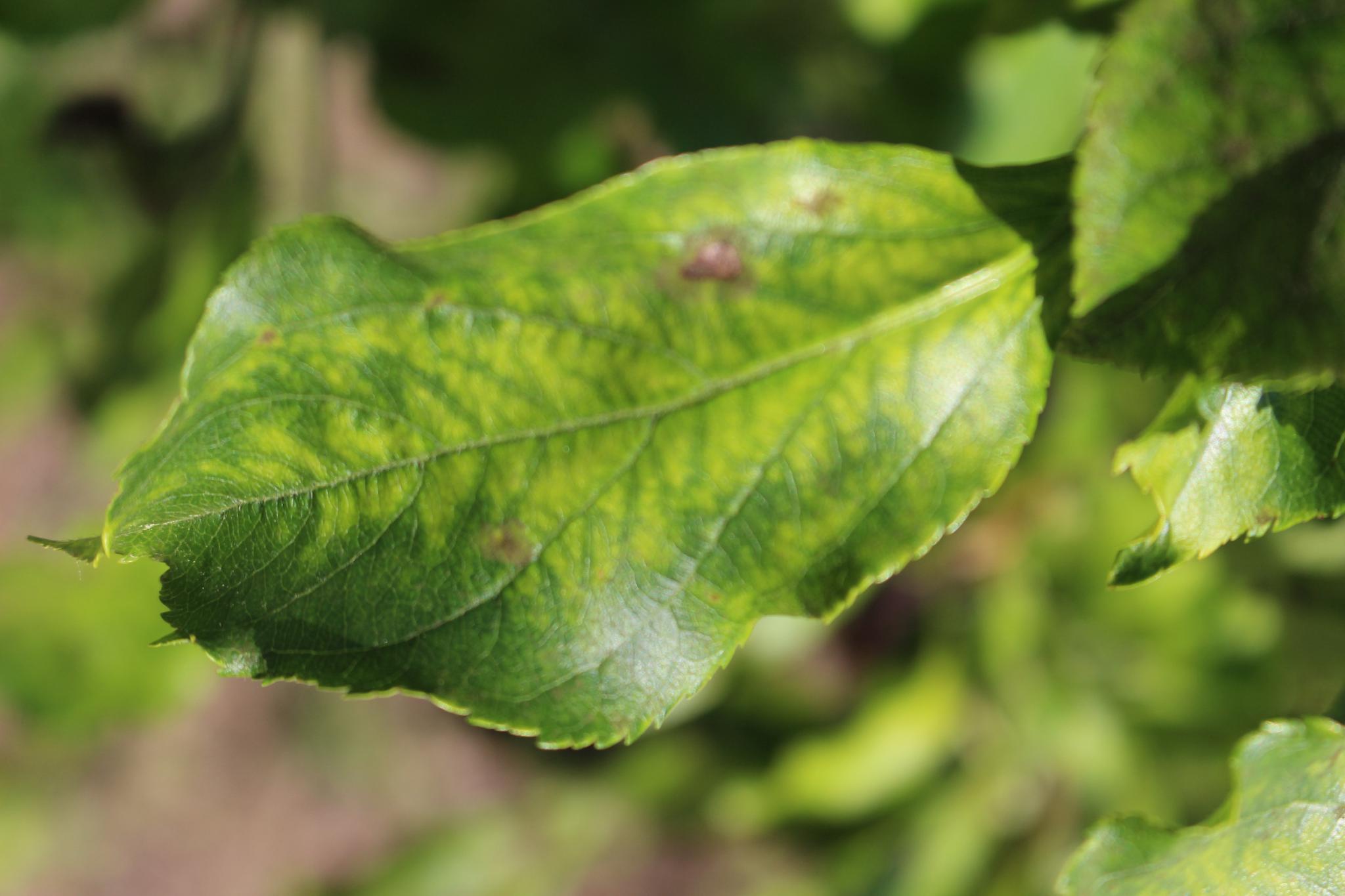}
    \caption{"Scab" Disease}
  \end{subfigure}
  \begin{subfigure}[b]{0.4\linewidth}
    \includegraphics[width=\linewidth]{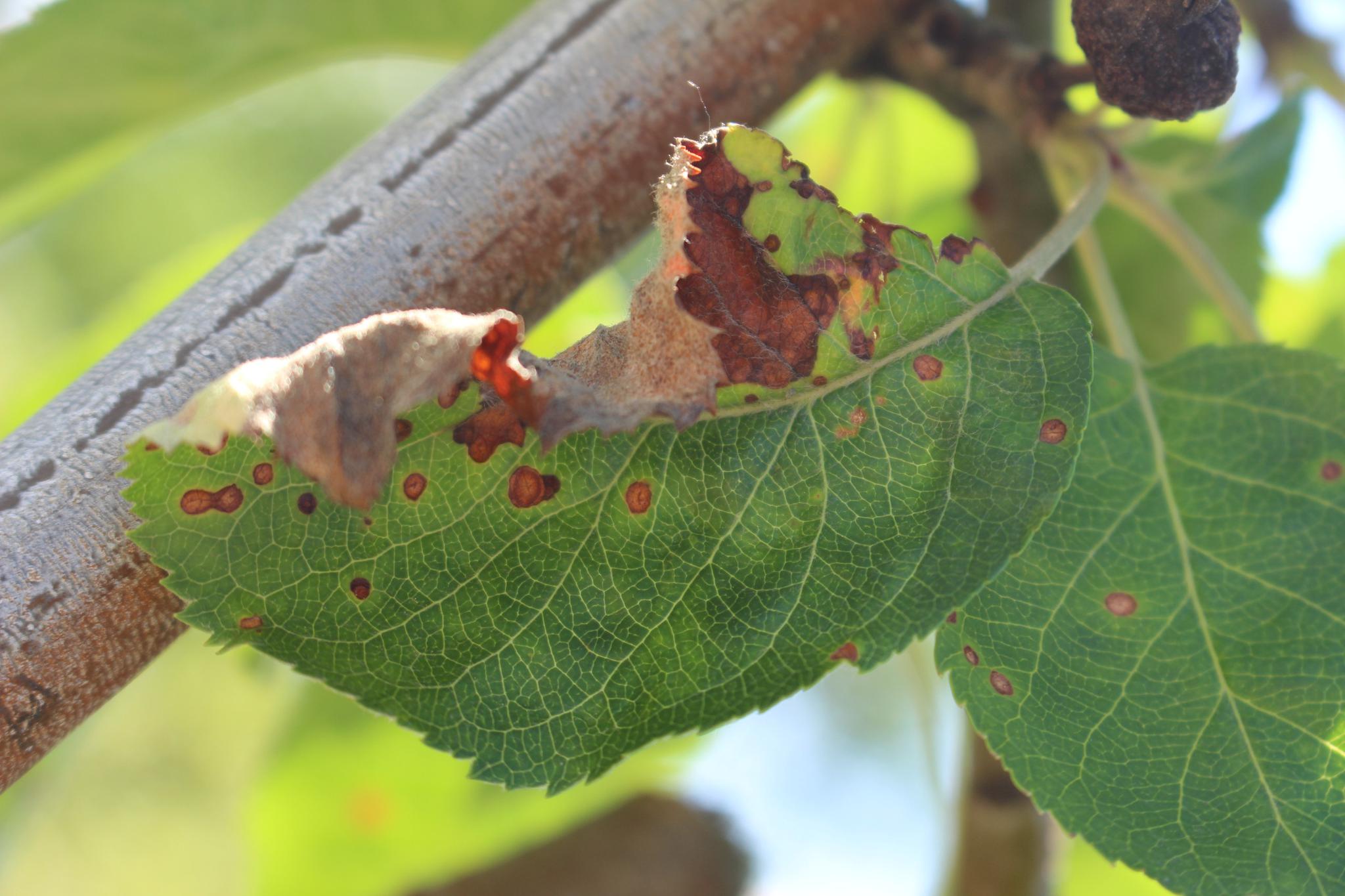}
    \caption{Both "Rust" \& "Scab"}
  \end{subfigure}
  \caption{Sample leaf images from the dataset}
  \label{fig:coffee}
\end{figure}

The target output prediction for each image is a vector of $4$ values, representing the probabilities of the leaf being in each of the possibilities (“healthy”, “multiple$\_$diseases”, “rust”, “scab”).

\subsection{Data Preprocessing}
Before training, some augmentations and preprocessing techniques are first performed on the images. All input images were first reshaped to have dimensions $(545 \times 545 \times 3)$.

For the training data, the input image is randomly flipped horizontally with a probability of $0.5$, as well as flipped vertically with a probability of $0.5$. Next, there is a random scalar shift and rotation, with rotation limit $25$ degrees and probability $0.7$. Furthermore, with a $0.5$ probability, we randomly select one of the following to apply to the images: 
\begin{itemize}
\item \textbf{Emboss} the image
\item \textbf{Sharpen} the image
\item Apply \textbf{blur} on the image
\end{itemize}
Finally, there is a piecewise affine shift with probability $0.5$, followed by channel-wise normalization. The purpose of these augmentations is to prevent overfitting and to add variations to make the training task "harder" so that the model can learn a better representation.

\begin{figure}[h!]
  \centering
  \begin{subfigure}[b]{0.9\linewidth}
    \includegraphics[width=\linewidth]{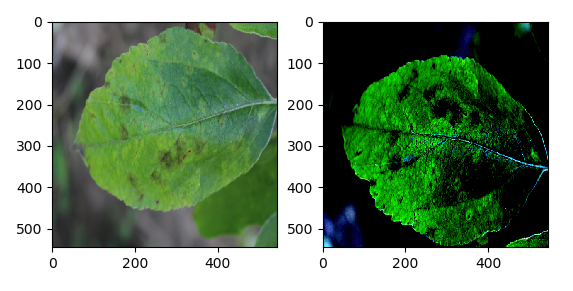}
  \end{subfigure}
  \caption{Preprocessing, before (left) and after (right)}
  \label{fig:coffee}
\end{figure}

These steps are \textbf{not} done for the validation and testing data. The only preprocessing done for validation and testing data is channel-wise normalization.

\section{Problem Statement}
The main problem is as follows: Given a $(545 \times 545 \times 3)$ image of a leaf, we want to identify whether or not the plant has a particular disease, and which specific disease the plant has. Specifically, our output will be four probabilities of the leaf being in each of our four classes.
\subsection{Loss Function}
During training, the loss function used is \textbf{cross-entropy}. This compares the similarity of two probability distributions and is given by the formula $$\textbf{Loss} = \sum_{i=1}^C p_i log(y_i)$$
Here, $C$ is the number of classes ($C=4$ in our case). $p_i$ represents the actual probabilities, and $y_i$ represents the predicted probabilities.
\subsection{Evaluation Metrics}
For evaluation, we follow the metric used by the official Kaggle competition, which is the mean column-wise ROC AUC. In other words, the score is the average of the individual AUCs of each predicted column. The ROC AUC is dependent on certain values such as the true positive rate (TPR) and false positive rate (FPR).

\section{Methodology}
\subsection{Benchmark Transfer Learning Models}
The idea of Transfer Learning is to use the weights of a large model pre-trained on another dataset and apply them on a novel task. Some of these large models include AlexNet \cite{alexnet}, VGGNet \cite{vgg}, ResNet \cite{resnet}, and more recently, DenseNet \cite{densenet} and EfficientNet \cite{efficientnet}. Since the early 2010s, one main dataset used to train and evaluate these large models is the ImageNet dataset \cite{imagenet}. Since then, the models have achieved progressively better performances on various tasks.

In this report, our main approach will be to begin with a baseline simple CNN model, followed by using these benchmark models. Eventually, we will progress to more state-of-the-art models such as EfficientNet and even semi-supervised Noisy Student traning. We will then compare the performance of various models and demonstrate that Noisy Student self-training model outperforms previous state-of-the-art models on this particular dataset.

\subsubsection{VGGNet}
The first model we attempt will be the VGGNet. The VGGNet is basically a stack of multiple CNN layers with dropout and pooling. Here, convolutional layers are of size $2 \times 2$ or $3 \times 3$, which is smaller compared to the previous AlexNet model. More specifically, we use the \textbf{VGG16}.
\begin{figure}[h!]
  \centering
  \begin{subfigure}[b]{0.9\linewidth}
    \includegraphics[width=\linewidth]{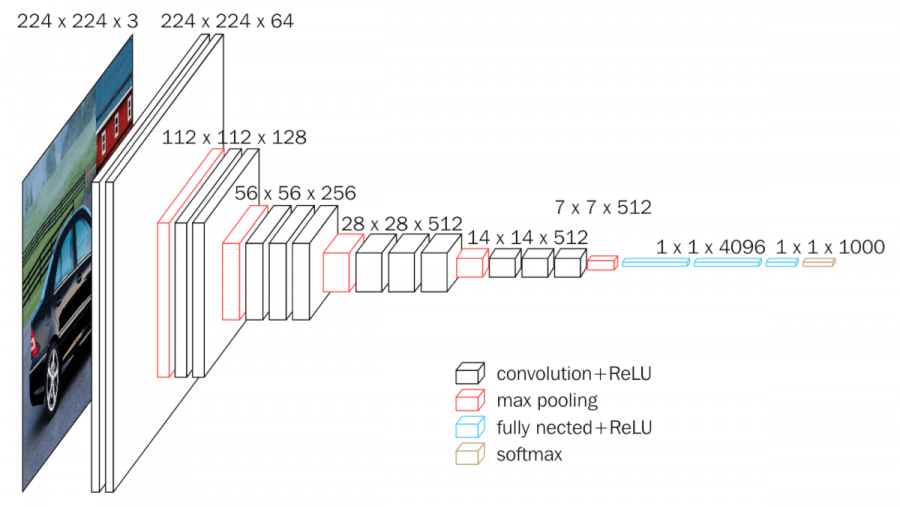}
  \end{subfigure}
  \caption{VGG16 Architecture}
  \label{fig:coffee}
\end{figure}

\subsubsection{ResNet}
After the VGGNet, the next model we try will be the ResNet \cite{resnet}. The ResNet introduces the new idea of "skip connections", as shown in the figure below. These special connections allow the gradients to be able to propagate backward more effectively, hence improving the performance.
\begin{figure}[h!]
  \centering
  \begin{subfigure}[b]{0.6\linewidth}
    \includegraphics[width=\linewidth]{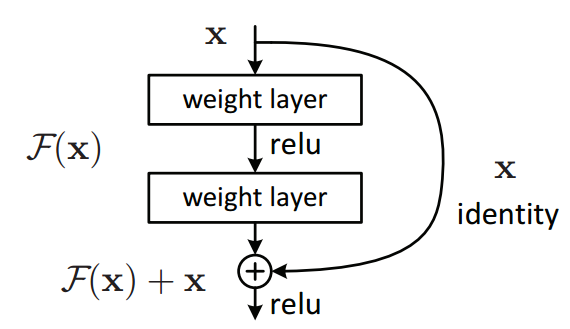}
  \end{subfigure}
  \caption{ResNet Skip Connection}
  \label{fig:coffee}
\end{figure}

More specifically, we use the \textbf{ResNet101} model.

\subsubsection{DenseNet}
After the ResNet, the next model we try on the dataset is the DenseNet \cite{densenet}. Here, the basic building block is called a dense block. It follows a similar idea to the ResNet, except that in the DenseNet, there are bottleneck connections across multiple dense blocks. This helps weights and gradients propagate even better.
\begin{figure}[h!]
  \centering
  \begin{subfigure}[b]{0.8\linewidth}
    \includegraphics[width=\linewidth]{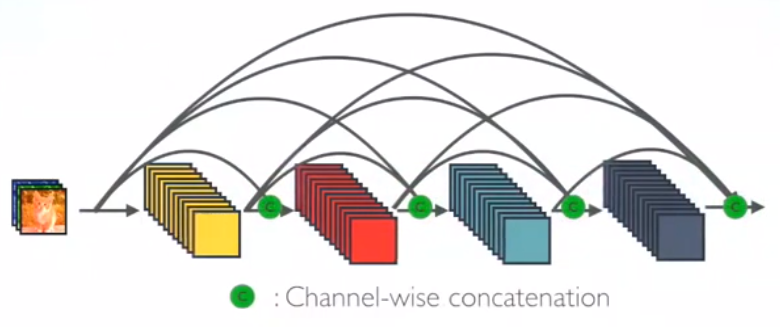}
  \end{subfigure}
  \caption{Dense Block Architecture}
  \label{fig:random}
\end{figure}

More specifically, we use the \textbf{DenseNet161} model.

\subsection{EfficientNet}
From the name, EfficientNet \cite{efficientnet}, released by Google last 2019, is not only about accuracy, but also about efficiency. It scales the model to perform much faster with less parameters than other previous SOTA models.

\begin{figure}[h!]
  \centering
  \begin{subfigure}[b]{0.9\linewidth}
    \includegraphics[width=\linewidth]{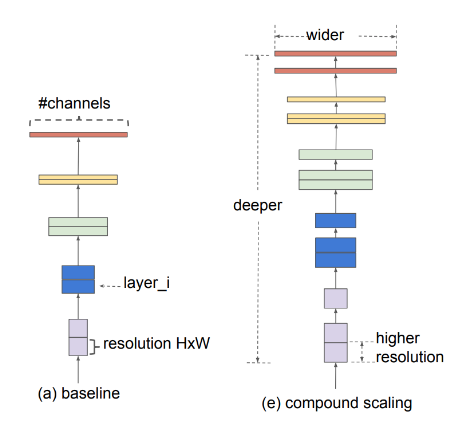}
  \end{subfigure}
  \caption{Combined Scaling for EfficientNet}
  \label{fig:scaleeffnet}
\end{figure}

EfficientNets use the idea of \textbf{model scaling} (see  Figure \ref{fig:scaleeffnet}). In any model, there are three types of scaling: 
\begin{enumerate}
\item \textbf{Depth Scaling}: This is the most common form of scaling. It basically means making a model deeper. In most cases, this will increase the accuracy of the model. However, as the model goes deeper, it may also lead to problems such as vanishing gradients, so the accuracy gains start to diminish if we go too deep.
\item \textbf{Width Scaling}: This refers to how wide the network is (for example, how many channels there are in a convolutional layer). The advantage of having a wider model is that it's smaller and able to capture fine-grained features. However, the disadvantage is that the accuracy will saturate quickly.
\item \textbf{Resolution Scaling}: This simply refers to the resolution of the input iamge. Inuitively, higher images will be better for the model, but practically, in high resolutions, there is no significant difference between, say, $500 \times 500$ as opposed to $560 \times 560$.
\end{enumerate}

In conclusion, we see that depth, width, or resolution scaling may improve performance, but the accuracy gains eventually diminish for larger models. 

\subsubsection{Combined Scaling}
EfficientNet uses the idea of \textbf{combined scaling}, which combines the three types of scaling to achieve optimal performance. It achieves this balance by considering the following constraints: 
$$\text{depth: } d = \alpha^\phi$$
$$\text{width: } w = \beta^\phi$$
$$\text{resolution: } r = \gamma^\phi$$
$$\text{s.t  } \quad \alpha \cdot \beta^2 \cdot \gamma^2 \approx 2 $$
$$\alpha \geq 1, \beta \geq 1, \gamma \geq 1$$
where $\phi$ is a user-specified parameter. These constraints are motivated by the computational idea that the FLOPS of a convolution operation is usually proportional to $d \cdot w^2 \cdot r^2$, so implementing these constraints effectively allows us to control the number of FLOPS.

\subsubsection{Network Structure}
For the building block of the EfficientNet, it actually uses MBConv layers, which are Inverted Residual Blocks that were introduced in MobileNet V2 \cite{mobilenet}. There are also some Squeeze and Excitation blocks \cite{squeeze}. To find the exact architecture, the authors employed a Neural Architecture Search (NAS) to find the optimal values of parameters $\alpha, \beta, \gamma, \phi$. Once these parameters are found, the full EfficientNet model can then be constructed. 

There are multiple configurations of EfficientNet. In this report, we use \textbf{Efficientnet B5}.

\subsection{Noisy Student Training}
Released in November 2019, \textbf{Noisy Student training} \cite{NoisyStudent} is the latest state-of-the-art in ImageNet classification. It is a semi-supervised self-training model that uses both labeled and unlabeled data. For the ImageNet task, the unlabeled dataset used is the JFT-300M datset \cite{jft}. One can clearly observe the massive scale of this unlabeled dataset (300M as compared to ImageNet's 14M). To handle the unlabeled data, the algorithm trains two types of networks: a teacher network and a student network.

\subsubsection{Teacher-Student Architecture}
The basic idea is to train a teacher model on labeled images to predict a distribution of class labels, then applying this to unlabeled data to create pseudo-labels. Then, a student model (usually larger) will be trained on both originally labeled data and the new pseudo-labeled data. This is done iteratively, meaning that the student model will then be used as a new teacher model to re-predict the pseudo-labels, and a new student model will be trained, and so on. Both student and teacher models are EfficientNet models, and the new student models usually become progressively larger as they scale up through the iterations.

\begin{figure}[h!]
  \centering
  \begin{subfigure}[b]{\linewidth}
    \includegraphics[width=\linewidth]{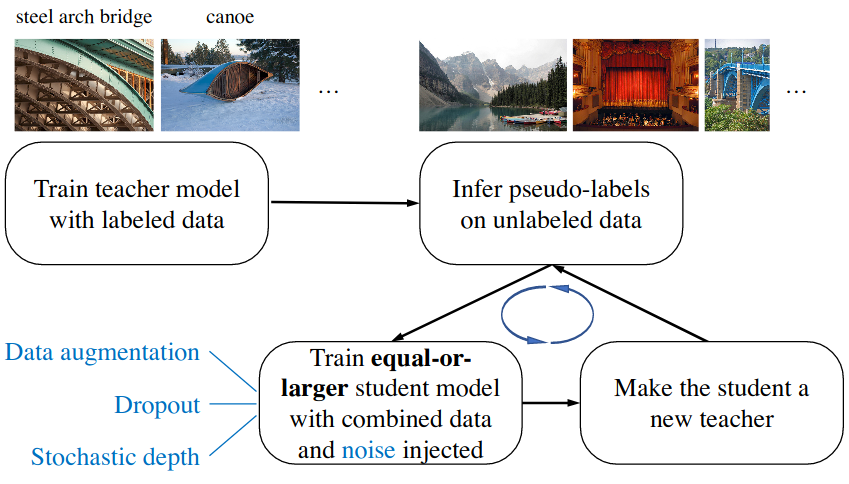}
  \end{subfigure}
  \caption{Noisy Student Pipeline}
  \label{fig:effnet}
\end{figure}

\subsubsection{Adding Noise to the Student Model}
The authors of the Noisy Student paper also discovered that adding a noise to the student networks was very significant to improving the performance. One particular useful form of noise in this model is Stochastic Depth \cite{stoch_depth}, which means adding random shortcuts between layers. Other forms of noise include data augmentation (rotation, horizontal and vertical shifts), and dropout \cite{dropout} (randomly discarding certain neurons). The main idea is to make the learning more difficult for the student model, which will allow it to learn more meaningful representations than the teacher, and therefore the model becomes more robust.

\section{Experiments}
In this report, we demonstrate the different experiments done on the various transfer learning models outlined in Section 5. Because these models were quite large, training was done with a batch size of \textbf{4} with the \textbf{Adam} optimizer \cite{adam} and a learning rate of \textbf{1e-3}. Training was done for \textbf{30 epochs} with a learning rate decay of \textbf{1e-3}, and the loss and accuracy were measured after every epoch. 

\subsection{Benchmark Models}
We first begin with a basic three-layer convolutional network. This is followed by our three benchmark transfer learning models, namely VGGNet, ResNet, and DenseNet. These models were pre-trained, and a linear classifier was added at the end. The results of the experiments are shown below:

\begin{table}[h!]
\begin{center}
\begin{tabular}{|l|l|ll}
\cline{1-2}
\textbf{Model} & \textbf{Test ROC AUC} &  &  \\ \cline{1-2}
3 Layer CNN    & 0.615                 &  &  \\ \cline{1-2}
VGG16          & 0.933                 &  &  \\ \cline{1-2}
ResNet101      & 0.943                 &  &  \\ \cline{1-2}
DenseNet161      & 0.945                 &  &  \\ \cline{1-2}
\end{tabular}
\end{center}
\caption{Test ROC AUC for the different models}
\end{table}

From the table above, we see that the three-layer CNN performs very poorly on the dataset, while the other three transfer learning models perform quite well on this dataset, with VGG performing significantly better than the simple three-layer CNN, then ResNet performing slightly better than VGGNet, and finally DenseNet achieving an ROC AUC score of around $0.945$. However, while these may seem quite high, it is actually not that high for this specific task. The DenseNet score of $0.945$ ranks around 850 out of 1345 in the Kaggle public leaderboard. We see later on that our EfficientNet model greatly outperforms this.

In order to further examine the DenseNet model and its common failures, we take the DenseNet and plot its confusion matrix. (See Figure \ref{fig:confmat_dense} in next page.)

\begin{figure}[h!]
  \centering
  \begin{subfigure}[b]{\linewidth}
    \includegraphics[width=\linewidth]{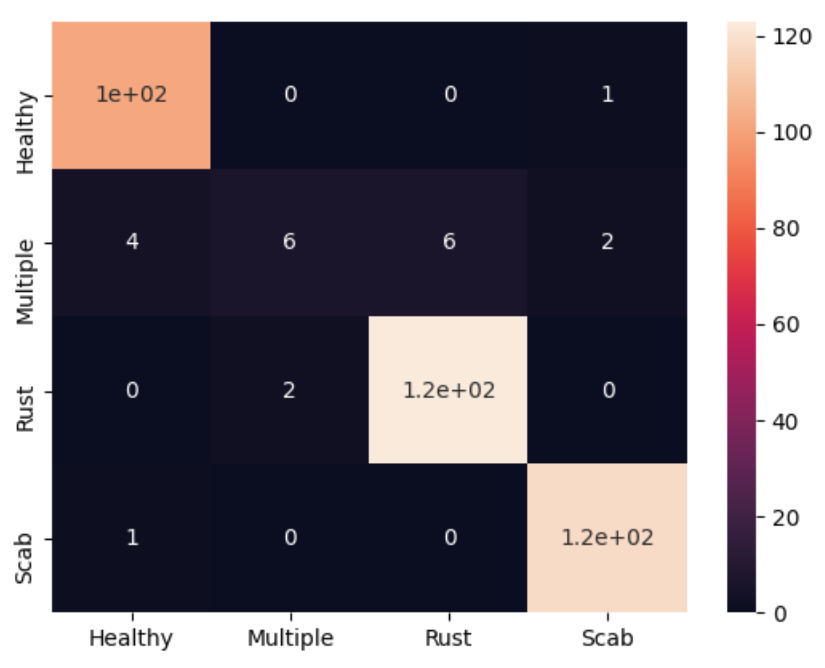}
  \end{subfigure}
  \caption{DenseNet Confusion Matrix}
  \label{fig:confmat_dense}
\end{figure}

From this, confusion matrix, we see that most of the errors were in the second row, i.e. when the leaf class was "multiple diseases". Only 6 images with multiple diseases were classified correctly, while 12 were classified incorrectly. These errors are consistent with our intuition, since we would expect that plants with multiple diseases would be harder to classify.

\subsection{EfficientNet Architectures}
EfficientNet architectures perform significantly better than the previous benchmark models. The baseline EfficientNet achieves a leaderboard test accuracy of $0.962$. Meanwhile, using weights from the Noisy Student Training achieves a leaderboard test accuracy of $0.967$. This is indeed a big jump compared to DenseNet at $0.945$

\begin{table}[h!]
\begin{center}
\begin{tabular}{|l|l|ll}
\cline{1-2}
\textbf{Model} & \textbf{Test ROC AUC} &  &  \\ \cline{1-2}
3 Layer CNN    & 0.615                 &  &  \\ \cline{1-2}
VGG16          & 0.933                 &  &  \\ \cline{1-2}
ResNet101      & 0.943                 &  &  \\ \cline{1-2}
DenseNet161      & 0.945                 &  &  \\ \cline{1-2}
EfficientNet      & \textbf{0.962}                 &  &  \\ \cline{1-2}
EfficientNet w/ Noisy Student      & \textbf{0.967}                 &  &  \\ \cline{1-2}
\end{tabular}
\end{center}
\caption{Summary of Final Results}
\end{table}

These observations are consistent with what we expect based on their performance on ImageNet and other similar tasks. The EfficientNet with Noisy Student performance is actually very good on this task. With an ROC AUC accuracy of $0.967$, this already outperforms most of the models in literature, and it achieves this without extensive feature engineering or model ensembling. 

\begin{figure*}
  \includegraphics[width=\textwidth,height=7cm]{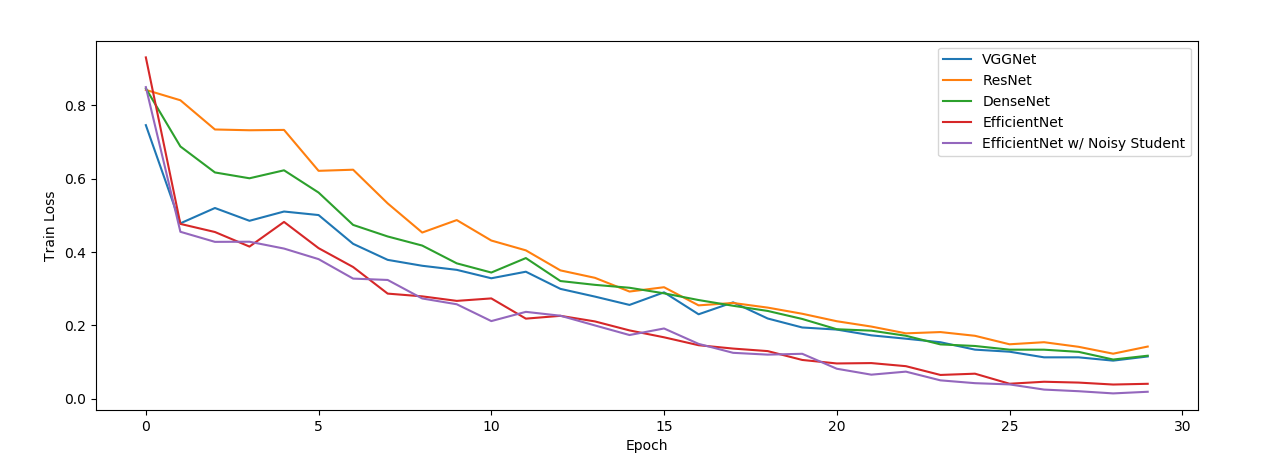}
  \includegraphics[width=\textwidth,height=7cm]{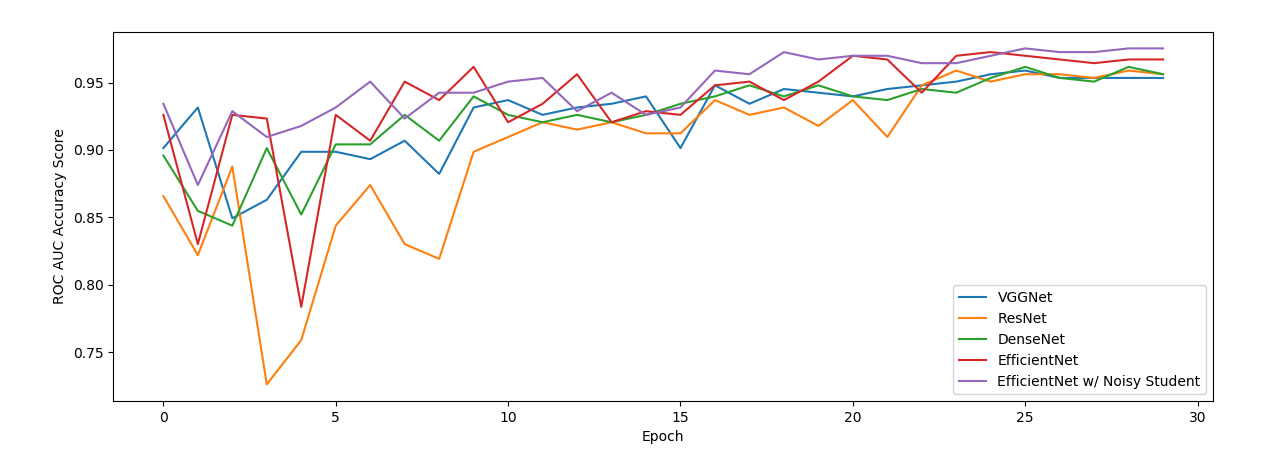}
  \caption{Train Loss (top) and Validation Accuracy (bottom)}
  \label{fig:curves}
\end{figure*}

\subsubsection{Loss Curve Analysis (Figure \ref{fig:curves})}
The improvement of EfficientNet models and Noisy Student Models can be seen most evidently in the loss and accuracy curves plotted at Figure  \ref{fig:curves} (see next page). In Figure \ref{fig:curves} in the next page, we have two plots, one for training loss across epochs, and another for validation ROC AUC accuracies.

In the curve of training losses, it is evident that EfficientNet and Noisy Student have consistently lower losses than the other models. These EfficientNet architectures also converge faster and are less prone to fluctuations. Comparing the baseline EfficientNet with the Noisy Student Training, we can also notice that using Noisy Student shows a small but clear improvement.

A similar pattern is shown in the curves for validation accuracies across epochs. We see that as they converge, the EfficientNet and Noisy Student models have a much higher validation accuracy as compared to VGGNet, ResNet, and DenseNet. Here, we also notice that there are many fluctuations in the early epochs for EfficientNet (and ResNet). However, once we add the Noisy Student weights, it becomes much more stable and converges faster, converging at around epoch 18 as opposed to the baseline EfficientNet, which conveges at around epoch 23.

\subsubsection{Confusion Matrix}
To further examine the performance of the EfficientNet with Noisy Student training, we visualize the confusion matrix, just as we did with the DenseNet.

\begin{figure}[h!]
  \centering
  \begin{subfigure}[b]{0.9\linewidth}
    \includegraphics[width=\linewidth]{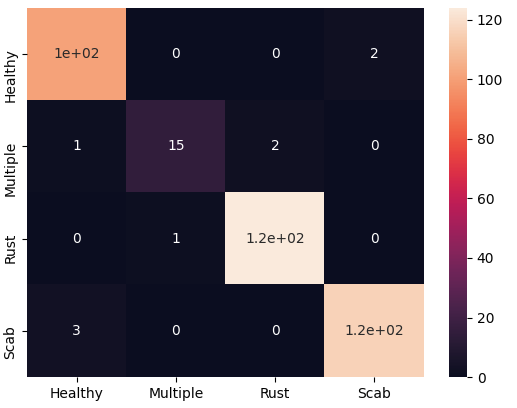}
  \end{subfigure}
  \caption{Noisy Student Training Confusion Matrix}
  \label{fig:confmat_dense}
\end{figure}

Indeed, we can see the confusion matrix of the Noisy Student no longer suffers from the same shortcomings of the DenseNet. The plants with multiple diseases are classified accurately. 

\subsection{Hyperparameter Tuning (For Noisy Student)}
Since we are using pre-trained models, there are not that many hyperparameters to tune. Furthermore, we are forced to use small batches due to GPU memory restrictions. Here, we focus on two main ones: \textbf{learning rate} and \textbf{optimizer}.

\subsubsection{Optimizer}
For the optimizers, we compare Adam and SGD
\begin{table}[h!]
\begin{center}
\begin{tabular}{|l|l|ll}
\cline{1-2}
\textbf{Optimizer} & \textbf{Test ROC AUC} &  &  \\ \cline{1-2}
SGD           & 0.9669                 &  &  \\ \cline{1-2}
Adam          & \textbf{0.9672}                 &  &  \\ \cline{1-2}
\end{tabular}
\end{center}
\caption{Test ROC AUC for the different models}
\end{table}

From Table 3, we see that the final results of Adam and SGD were more or less the same. However, we noticed during training that Adam converged faster, so we select Adam as our optimizer.

\subsubsection{Learning Rate}
Below are the performances of different learning rates.
\begin{table}[h!]
\begin{center}
\begin{tabular}{|l|l|ll}
\cline{1-2}
\textbf{Learning Rate} & \textbf{Test ROC AUC} &  &  \\ \cline{1-2}
1e-1      & 0.9437                 &  &  \\ \cline{1-2}
1e-2      & 0.9653                 &  &  \\ \cline{1-2}
1e-3      & \textbf{0.9672}                 &  &  \\ \cline{1-2}
1e-4      & 0.9667                 &  &  \\ \cline{1-2}
\end{tabular}
\end{center}
\caption{Performance of different learning rates}
\end{table}

We see in Table 4 that the ROC AUC for various learning rates are more or less the same (except for 1e-1, which is much lower). However, the performance for 1e-3 is slightly better than the rest, so that is the one we select.

\subsection{Additional Techniques}
\subsubsection{Model Ensembling}
Model Ensembling is the idea of combining the losses of multiple models into one so that the models can compensate for the shortcomings of the other models. Here, I combined my best EfficientNet model with my two best Noisy Student models.

\begin{figure}[h!]
  \centering
  \begin{subfigure}[b]{\linewidth}
    \includegraphics[width=\linewidth]{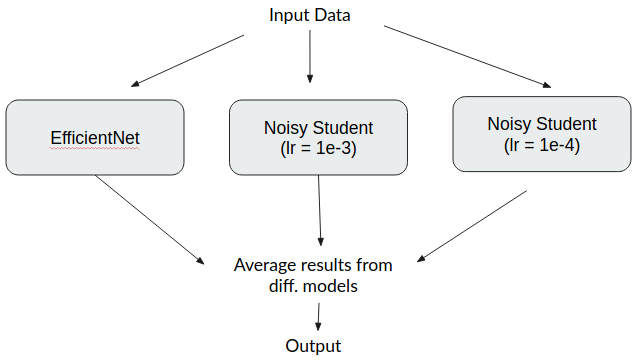}
  \end{subfigure}
  \caption{Model Ensembling}
  \label{fig:confmat_dense}
\end{figure}

The final ensembled model had a test score of \textbf{0.982} and ranked 64 out of 1345 (top 5$\%$) on the public leaderboard.

\section{Conclusion}
In this report, we applied various CNN-based image classification transfer learning models to the novel Plant Pathology Classification task. We discovered that the EfficientNet model outperforms the other models significantly. Additionally, we see that adding Noisy Student Training to the EfficientNet model further improves the model performance and makes the model converge faster and less prone to fluctuations.

\section{Future Work}
The field of plant pathology classification is a very significant one with large global impacts. This paper showed the use of EfficientNet architectures on apple leaves. In the future, more tests can be done on various other types of plants and fruits. Since the accuracy for this task is already quite high, one possible area of improvement would be the speed of detection and classification (real-time detection, mobile devices, etc.)

{\small
\bibliographystyle{ieee}
\bibliography{egbib}

\begin{thebibliography}{10}\itemsep=-1pt

\bibitem{good}
K.~R. Aravind and P.~Raja.
\newblock Automated disease classification in (selected) agricultural crops
  using transfer learning.
\newblock {\em Automatika}, 61(2):260--272, 2020.

\bibitem{imagenet}
J.~Deng, W.~Dong, R.~Socher, L.-J. Li, K.~Li, and L.~Fei-Fei.
\newblock {ImageNet: A Large-Scale Hierarchical Image Database}.
\newblock In {\em CVPR09}, 2009.

\bibitem{resnet}
K.~He, X.~Zhang, S.~Ren, and J.~Sun.
\newblock Deep residual learning for image recognition.
\newblock {\em CoRR}, abs/1512.03385, 2015.

\bibitem{squeeze}
J.~Hu, L.~Shen, and G.~Sun.
\newblock Squeeze-and-excitation networks.
\newblock {\em CoRR}, abs/1709.01507, 2017.

\bibitem{densenet}
G.~Huang, Z.~Liu, and K.~Q. Weinberger.
\newblock Densely connected convolutional networks.
\newblock {\em CoRR}, abs/1608.06993, 2016.

\bibitem{stoch_depth}
G.~Huang, Y.~Sun, Z.~Liu, D.~Sedra, and K.~Q. Weinberger.
\newblock Deep networks with stochastic depth.
\newblock {\em CoRR}, abs/1603.09382, 2016.

\bibitem{Khirade}
S.~D. {Khirade} and A.~B. {Patil}.
\newblock Plant disease detection using image processing.
\newblock pages 768--771, 2015.

\bibitem{adam}
D.~Kingma and J.~Ba.
\newblock Adam: A method for stochastic optimization.
\newblock {\em International Conference on Learning Representations}, 12 2014.

\bibitem{alexnet}
A.~Krizhevsky, I.~Sutskever, and G.~E. Hinton.
\newblock Imagenet classification with deep convolutional neural networks.
\newblock In {\em Proceedings of the 25th International Conference on Neural
  Information Processing Systems - Volume 1}, NIPS’12, page 1097–1105, Red
  Hook, NY, USA, 2012. Curran Associates Inc.

\bibitem{clustering2}
C.~U. {Kumari}, S.~{Jeevan Prasad}, and G.~{Mounika}.
\newblock Leaf disease detection: Feature extraction with k-means clustering
  and classification with ann.
\newblock In {\em 2019 3rd International Conference on Computing Methodologies
  and Communication (ICCMC)}, pages 1095--1098, 2019.

\bibitem{Clustering}
S.~Maity, S.~Sarkar, A.~Tapadar, A.~Dutta, S.~Biswas, S.~Nayek, and P.~Saha.
\newblock Fault area detection in leaf diseases using k-means clustering.
\newblock 10 2018.

\bibitem{Survey}
A.~Patel and B.~Joshi.
\newblock A survey on the plant leaf disease detection techniques.
\newblock {\em IJARCCE}, 6:229--231, 01 2017.

\bibitem{Radha}
S.~Radha.
\newblock Leaf disease detection using image processing.
\newblock {\em Journal of Chemical and Pharmaceutical Sciences}, 03 2017.

\bibitem{rumpf}
T.~Rumpf, A.-K. Mahlein, U.~Steiner, E.-C. Oerke, H.-W. Dehne, and L.~Plümer.
\newblock Early detection and classification of plant diseases with support
  vector machines based on hyperspectral reflectance.
\newblock {\em Computers and Electronics in Agriculture}, 74:91--99, 10 2010.

\bibitem{mobilenet}
M.~Sandler, A.~G. Howard, M.~Zhu, A.~Zhmoginov, and L.~Chen.
\newblock Inverted residuals and linear bottlenecks: Mobile networks for
  classification, detection and segmentation.
\newblock {\em CoRR}, abs/1801.04381, 2018.

\bibitem{rice}
V.~K. {Shrivastava}, M.~K. {Pradhan}, S.~{Minz}, and M.~P. {Thakur}.
\newblock {Rice Plant Disease Classification Using Transfer Learning of Deep
  Convolution Neural Network}.
\newblock {\em ISPRS - International Archives of the Photogrammetry, Remote
  Sensing and Spatial Information Sciences}, 423:631--635, July 2019.

\bibitem{vgg}
K.~Simonyan and A.~Zisserman.
\newblock Very deep convolutional networks for large-scale image recognition.
\newblock {\em CoRR}, abs/1409.1556, 2014.

\bibitem{Singh}
V.~Singh and A.~Misra.
\newblock Detection of plant leaf diseases using image segmentation and soft
  computing techniques.
\newblock {\em Information Processing in Agriculture}, 4(1):41 -- 49, 2017.

\bibitem{dropout}
N.~Srivastava, G.~Hinton, A.~Krizhevsky, I.~Sutskever, and R.~Salakhutdinov.
\newblock Dropout: A simple way to prevent neural networks from overfitting.
\newblock {\em J. Mach. Learn. Res.}, 15(1):1929–1958, Jan. 2014.

\bibitem{jft}
C.~Sun, A.~Shrivastava, S.~Singh, and A.~Gupta.
\newblock Revisiting unreasonable effectiveness of data in deep learning era.
\newblock {\em CoRR}, abs/1707.02968, 2017.

\bibitem{efficientnet}
M.~Tan and Q.~V. Le.
\newblock Efficientnet: Rethinking model scaling for convolutional neural
  networks, 2019.
\newblock cite arxiv:1905.11946Comment: Published in ICML 2019.

\bibitem{dataset}
R.~Thapa, N.~Snavely, S.~Belongie, and A.~Khan.
\newblock The plant pathology 2020 challenge dataset to classify foliar disease
  of apples, 2020.

\bibitem{Toda}
Y.~{Toda} and F.~{Okura}.
\newblock How convolutional neural networks diagnose plant disease.
\newblock {\em Plant Phenomics}, 2019.

\bibitem{NoisyStudent}
Q.~Xie, E.~Hovy, M.-T. Luong, and Q.~Le.
\newblock Self-training with noisy student improves imagenet classification.
\newblock 11 2019.

\end{thebibliography}
}

\end{document}